\pdfoutput=1

\documentclass[11pt]{article}
\usepackage[preprint]{acl}

\usepackage{acl}

\usepackage{xcolor}
\usepackage{adjustbox}
\usepackage[most]{tcolorbox}
\usepackage{multirow}
\usepackage{booktabs}
\usepackage{colortbl}
\usepackage{times}
\usepackage{inconsolata}
\usepackage{latexsym}
\usepackage{multirow}
\usepackage{graphicx}
\usepackage{subcaption} 
\usepackage{algorithm,algpseudocode}
\usepackage[inline]{enumitem}
\usepackage{
  amsfonts, nicefrac, amsmath, bm,
  soul, graphicx, multirow, wrapfig, lipsum, booktabs, caption, subcaption, stfloats, listings,
  longtable, paralist, cleveref, amssymb, pifont
}
\usepackage{enumitem}

\definecolor{promptblue}{RGB}{181, 179, 242}

\newtcolorbox[list inside=prompt,auto counter,number within=section]{prompt}[1][]{
    colbacktitle=black!60,
    fonttitle=\small,
    coltitle=white,
    fontupper=\footnotesize,
    boxsep=4pt,
    left=0pt,
    right=0pt,
    top=0pt,
    bottom=0pt,
    boxrule=1pt,
    #1,
    breakable,
}

\definecolor{britishracinggreen}{rgb}{0.0, 0.26, 0.15}

\definecolor{lightorange}{RGB}{255, 234, 222}
\definecolor{burntorange}{RGB}{191, 97, 34}
\definecolor{coolblue}{RGB}{181, 179, 242}
\definecolor{lightblue}{RGB}{195, 224, 247}
\definecolor{darkblue}{RGB}{79, 118, 148}

\newcommand{\gpt}{\texttt{gpt-4o}}
\newcommand{\deepseek}{\texttt{deepseek-v3}}
\newcommand{\gptmini}{\texttt{gpt-4o-mini}}
\newcommand{\llama}{\texttt{llama-3-8B-Instruct}}
\newcommand{\llamashortened}{\texttt{llama-3-8B}}

\newcommand{\minstral}{\texttt{Ministral-3B}}
\newcommand{\llamabig}{\texttt{llama-3-70B-Instruct}}

\newcommand{\rewriter}{\texttt{rewriter}}
\newcommand{\chatbot}{\texttt{chatbot}}

\newcommand{\ignore}[1]{}

\usepackage[T1]{fontenc}

\usepackage[utf8]{inputenc}
\usepackage{lineno}
\usepackage{microtype}

\usepackage{mparhack}

\definecolor{MutedOrange}{RGB}{255, 235, 200}
\definecolor{MutedBlue}{RGB}{200, 200, 255}

\usepackage{hyperref}
\hypersetup{
citecolor = [RGB]{59,52,145},
linkcolor = [RGB]{59,52,145},
urlcolor = [RGB]{59,52,145}
}

\crefname{section}{\S}{\S\S}

\title{Conversational User-AI Intervention: A Study on Prompt Rewriting for Improved LLM Response Generation}
\author{
  \textbf{Rupak Sarkar\textsuperscript{$^{2}$}}\thanks{\small{Work done as an intern at Microsoft Research. Corresponding authors: rupak@umd.edu, sjauhar@microsoft.com.}},
  \textbf{Bahareh Sarrafzadeh\textsuperscript{$^{1}$}},
  \textbf{Nirupama Chandrasekaran\textsuperscript{$^{1}$}}, \\
  \textbf{Nagu Rangan\textsuperscript{$^{1}$}},
  \textbf{Philip Resnik\textsuperscript{$^{2}$}},
  \textbf{Longqi Yang\textsuperscript{$^{1}$}},
  \textbf{Sujay Kumar Jauhar\textsuperscript{$^{1}$}}
  \\
  \\
  \textsuperscript{$^{1}$}Microsoft Corporation, Redmond;
  \textsuperscript{$^{2}$}University of Maryland, College Park
}

\begin{document}
\maketitle

\begin{abstract}
Human-LLM conversations are increasingly pervasive in peoples' professional and personal lives, yet many users still struggle to elicit helpful responses from LLM chatbots. 
This issue stems partly from users' lack of understanding in crafting effective prompts that accurately convey their information needs. 
Real-world conversational datasets combined with the text understanding faculties of LLMs present a unique opportunity to study this problem, and its potential solutions, at scale.
We present a large-scale LLM-centric study of real human-AI conversations, examining how user queries fall short of expressing information needs and exploring the use of LLMs to rewrite suboptimal user prompts.
Our findings show that rephrasing ineffective prompts can elicit better responses from a conversational system, while preserving user intent.
Notably, rewrites become more effective in longer conversations where contextual inferences about user needs can be made more accurately.
We observe that LLMs often need to---and naturally do---make \emph{plausible} assumptions about user intentions and goals when interpreting prompts. 
These findings largely hold across conversational domains, user intents, and various LLM sizes and families, suggesting prompt rewriting as a promising approach for improving human-AI interactions.
\end{abstract}

\section{Introduction}\label{sec:intro}
Many technologies we have come to rely on in our daily lives---from search engines to cellphones---now feature LLM chat interfaces.
These powerful models have unlocked new frontiers in conversational agents and automated reasoning~\citep{liu2024llmconversationalagentmemory, tian-etal-2024-large-language}.
However, users often struggle to obtain satisfactory responses from these systems~\cite{wang2024understandinguserexperiencelarge} or understand how their prompts influence LLM output~\cite{khurana2024iui}. 
A recent study of students writing code-generation prompts with an LLM assistant revealed that the success of a prompt largely depended on chance---though some prompts proved effective for certain models, students with similar Python expertise found it challenging to write prompts that worked consistently~\cite{babe-etal-2024-studenteval}. 

User queries may receive unsatisfactory responses for various reasons. 
Responses might be incorrect, irrelevant, or contain fabrications~\cite{li-etal-2024-dawn, yehuda-etal-2024-interrogatellm, xu2024hallucination}.
Other failures stem from unrealistic user expectations about AI system capabilities.
For example, users may not realize that ChatGPT cannot perform certain actions like taking screenshots.
Users may also be dissatisfied when AI systems abstain from answering queries that violate their guidelines (for example, ``watch Wicked online for free'').

Existing prompt design tools primarily target professionals and NLP practitioners~\cite{perreira-2023,schnabel2024prompts}, while most commercial chatbot users are laypeople without an intuitive understanding of crafting effective prompts. 
In fact, \citet{pooledayan2024} finds that undesirable LLM behavior disproportionately affects users with lower English proficiency and education. 

For equity and broader user engagement, it is imperative that LLMs deployed as chatbots better interpret users' information needs in whatever form they are expressed.
Understanding user-AI interaction failure at scale and testing remediation strategies is the first step towards developing better solutions. 
Publicly available datasets of human-LLM conversations, such as WildChat~\cite{zhao2024wildchat}, combined with the capability of modern LLM systems to analyze, interpret, and rewrite text at scale presents an opportunity to improve user-AI interactions.
While prior work has leveraged conversational logs to measure user satisfaction~\cite{lin2024interpretable}, generate taxonomies~\cite{wan2024tnt}, and perform alignment~\cite{shi2024wildfeedback}, no research effort to the best of our knowledge has examined how sub-optimal prompts impact unsatisfactory user outcomes.

We investigate whether LLMs can rewrite user prompts to \textbf{better express their information needs} and measure how these rewrites impact LLM-generated response quality.
Our investigative framework involves two LLMs: the \chatbot, which converses with the user, and the \rewriter, which rewrites prompts.
Given a conversational history between the user and \chatbot, we study whether the \rewriter \ can infer user information needs and produce improved prompts. Then, we measure the downstream impact of these rewrites by prompting the \chatbot~to respond to the reformulated prompt. 

During prompt rewriting, we also ask the \rewriter~to generate additional insights. These include the degree of modification required, the aspects of improvement (such as clarity, or specificity), and the assumptions, if any, the model needs to make in order to construct an effective prompt.\footnote{Sometimes, user prompts are so underspecified that it is impossible to infer underlying needs without making assumptions.}
Not only do these insights serve as a chain-of-thought for the model as it reformulates a user prompt, but they also provide novel axes along which to analyze user-AI conversations, and the impact of performing strategic interventions.

We apply our framework to WildChat conversations with unsatisfactory user outcomes.
Across five pairs of conversational domains and user intents and five different LLMs (various sizes, both open- and closed-source), we demonstrate that LLMs---including smaller ones---effectively rewrite prompts, producing consistently better chatbot responses.
Our experimental results are consistent with both \gpt \ and \deepseek \ as automatic evaluators, as well as with human judges. 
We also find that longer conversations yield better prompt rewrites, aspects of improvement partially overlap across domains, and that models make \textit{plausible} assumptions during rewriting.
\section{Preliminaries}\label{sec:prelims}

We begin by formalizing our problem space and the dataset we use for our framework.

\subsection{Problem Setup}\label{subsec:setup}

To study whether contextually intervening to rewrite human prompts with LLMs can improve response quality, we simulate its effectiveness retroactively.\footnote{The ideal setup to test the helpfulness of interventions would require in-situ A/B testing, which is beyond the scope of our work.} 
That is, we analyze real-world historical human-LLM conversations and rewrite user prompts at key turns, evidencing user dissatisfaction to show that those rewrites result in better responses. 
Admittedly, user dissatisfaction does not uniquely stem from sub-optimal prompts; unfounded user expectations, poor LLM responses, and abstentions due to safety policies could all be contributing factors. 
Therefore, we design our framework to be robust to these different types of conversational outcomes: rewrites should help with conversations that benefit from prompt reformulation, while broadly maintaining intent and not degrading performance on other types of conversations.\footnote{Our results (Section~\ref{sec:results}) demonstrate the general success of this approach, and we discuss other cases in Section~\ref{sec:insights}}

\begin{figure}[t]
\centering
\includegraphics[scale=0.6]{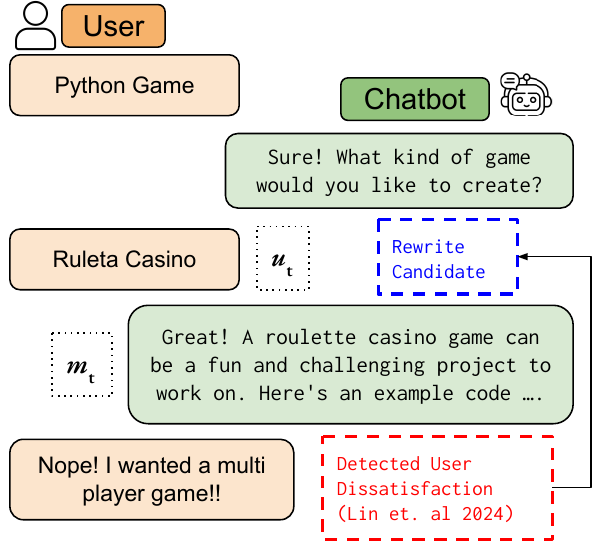}
\caption{\small{Figure showing candidate turn selection. We work backward from a user response expressing dissatisfaction to the query that caused the most recent model response.}}
\label{fig:dsat}
\end{figure}

Our problem setup is illustrated in Figure~\ref{fig:dsat}. Broadly, working backwards from a user turn that evidences dissatisfaction (\emph{DSAT}), we use a \rewriter\ LLM to reformulate the preceding user turn, and measure how a \chatbot\ LLM responds to this rewritten prompt. In Figure~\ref{fig:dsat} a vaguely written user prompt (``Ruleta Casino'') becomes a candidate for a prompt rewrite since it results in a response that dissatisfies the user.

More formally, consider a conversation $C = \{u_1,m_1,...u_n,m_n\}$ 
%
consisting of alternating user turns $u_i$ and model responses $m_i$, where $i$ is the index of a dialog turn. 
Moreover, define \chatbot~ to be an autoregressive LLM that responds to a user input at turn $i$: $m_i = LLM_{chatbot}(u_i, H_i; \theta)$, 
where $\theta$ are the set of model parameters, $H_i = (u_1, m_1, \dots, u_{i-1}, m_{i-1})$ refers to the conversational history up to user turn $u_i$. 
Also, assume \rewriter~ to be a (potentially different) LLM that rewrites an existing user prompt $u_i$: $u_{i}^{\prime} = LLM_{rewriter}(u_i, H_i, P; \theta^{\prime})$, 
where $P$ is a prompt template (Prompt~\ref{prompt:rewrite-prompt} in Appendix) the model uses to perform the rewrite.

Then, given a turn $u_d$ in a conversation $C$ that shows evidence of \emph{DSAT}, our problem becomes that of generating:
\begin{align}
u_{d-1}^{\prime} &= LLM_{rewriter}(u_{d-1}, H_{d-1}, P; \theta^{\prime}) \label{eq:main1} \\
m_{d-1}^{\prime} &= LLM_{chatbot}(u_{d-1}^{\prime}, H_i; \theta)
\label{eq:main2}
\end{align}
such that $Q(m_{d-1}^{\prime}) \geq Q(m_{d-1})$ by some quality measure $Q$.

\subsection{Dataset}\label{subsec:dataset}

In order to study this retroactive rewrite setup, we need---(a) A corpus of real-world user-LLM conversations; and (b) Labels indicating which turns result in user dissatisfaction.
For (a), we use a subset of WildChat consisting of non-toxic English conversations that have three or more turns. 
We follow the data setup of~\citet{shi2024wildfeedback}, who previously leveraged this subset.
Meanwhile, for (b) we use the user satisfaction rubrics proposed by~\citet{lin2024interpretable} and adapted by~\citet{shi2024wildfeedback} to assign a label of \emph{SAT}, \emph{DSAT} or \emph{NONE} to every turn in our dataset, and retain those conversations with at least on \emph{DSAT} label.

Our sample of the WildChat dataset is still large, and comprises chat interactions covering a wide variety of conversational domains and expressing a range of user intents. 
To further focus our study of how user dissatisfaction varies across these axes, we classify conversation turns into domains and intents~\cite{wan2024tnt}. Domains cover topical categories such as ``Software and Web Development'' and ``Culture and History'', while intents refer to the user's conversational goals such as ``seeking information'' and ``creation''.

\begin{table}[t]
    \centering
    \small
    \setlength{\tabcolsep}{3pt} 
    \resizebox{.95\columnwidth}{!}{
        \begin{tabular}{l l | ccc}
            \toprule
            \textbf{Domain} & \textbf{Intent} & \textbf{\#Convs} & \textbf{\#>=5 Turns} & \textbf{Rewrite Index} \\
            \midrule
            Software/Web Dev & Seek Info & 2397 & 446 & 2.71 \\
            
            Software/Web & Create & 1459 & 197 & 2.22 \\
            
            Writing/Journalism    & Create  & 376  & 64  & 2.39 \\
            
            Tech   & Seek Info & 349  & 78  & 3.49 \\
            
            Math/Logic    & Seek Info & 346  & 79  & 3.30 \\
            \bottomrule
        \end{tabular}
    }
    \caption{\small{Conversation metrics broken down by Domain and Intent. Rewrite index refers to the average turn corresponding to a candidate rewrite.}}
    \label{tab:domain_intent_stats}
\end{table}

In our analyses, we group conversations jointly over domain and intent to categorize them with similar information goals. 
We perform all our analyses for the five most commonly-occurring categories in our dataset, which can be found in Table~\ref{tab:domain_intent_stats}.

\section{Conversational Intervention through Prompt Rewriting}\label{sec:approach}

\begin{figure*}[t!]
\centering
\includegraphics[width=0.8\textwidth]{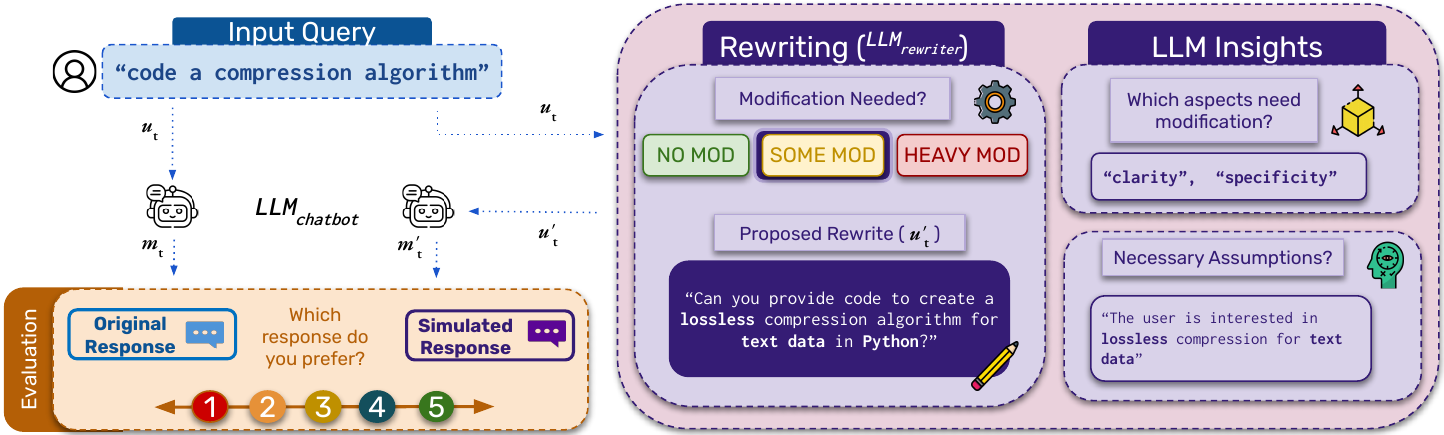}
\caption{\small{This figure summarizes our overall approach to prompt rewriting and evaluation. First, the \rewriter~processes an input user prompt along with the conversational history $H$ with the~\chatbot, and reasons about the aspects in which the query needs to improve, as well as the assumptions needed to make a rewrite before proposing a rewrite (Section~\ref{sec:approach}).
Then we comparatively measure the quality of the response to this rewritten prompt against the original, using either an LLM or a human as a judge (Section~\ref{sec:results}).}}
\vspace*{-3pt}
\label{fig:pipeline}
\end{figure*}

Given the cost of fine-tuning LLMs and the lack of relevant prompt rewriting data, we instead use a prompting strategy to perform rewrites. 
Broadly, our approach consists of two instruction categories: the first deals directly with the model performing the rewrite, while the second seeks to generate additional insights that serve as a chain-of-thought~\cite{wei2022chain} for better reasoning, as well as novel axes of analysis in our investigative framework. 
These two categories are illustrated in Figure~\ref{fig:pipeline} and are detailed below in Sections~\ref{subsec:rewrites} and~\ref{subsec:insights}.
Our full prompt is included in Appendix~\ref{prompt:rewrite-prompt}.

\subsection{Performing Rewrites}\label{subsec:rewrites}

Given a candidate turn $u_t$ we first instruct models to reason about the degree which it needs a rewrite on a 3-point scale -- \texttt{NO MOD} indicating that the \rewriter~has judged the prompt to be adequate, \texttt{SOME MOD}, and \texttt{HEAVY MOD}. This is to make our pipeline robust to cases that would otherwise not benefit from prompt rewriting.

Then for the cases identified as \texttt{SOME MOD} or \texttt{HEAVY MOD} we instruct models to generate a better, rewritten prompt $u_t^{\prime}$ while \emph{maintaining user intent}, according to Equation~\ref{eq:main1}. Using this rewritten prompt, we can then generate a new candidate \chatbot\ response $m_t^{\prime}$ using Equation~\ref{eq:main2}.

\subsection{Generating Additional Insights}\label{subsec:insights}

In addition to reformulating a prompt, we instruct models to perform fine-grained reasoning on additional insights about the rewriting operation.

\paragraph{Aspects of query improvement.}
The first category of insights are \emph{aspects}---open-ended free-text categories such as clarity, specificity, tone, etc. that models are instructed to list as they perform a rewrite (see Figure~\ref{fig:pipeline}).
We use these aspects to understand along what dimensions sub-optimal user queries fall short, as well as the ways LLMs perform rewriting operations.

\paragraph{Gathering Model Assumptions}
The second category of insights focuses on \emph{assumptions} that the model needs to make to effectively rewrite a prompt. 
This becomes especially relevant in cases when the user input is underspecified, or the historical conversation context lacks sufficient grounding information. In such cases, we instruct the model to list \emph{plausible} assumptions about the user's information goals.\footnote{\citet{hoyle-etal-2023-natural} showed that its possible to extract such \emph{plausible} inferences using LLMs.}
Figure~\ref{fig:pipeline} contains an example of an assumption the model makes while rewriting the input query.
Assumptions are useful for understanding how models reason about user needs, as well as measuring the impact that these assumptions have on LLM response quality.
\section{Results}\label{sec:results}

Before presenting our experiments and results, we summarize how pairs of original and candidate responses $m_t$ and $m_t^{\prime}$ (from Equations~\ref{eq:main1} and~\ref{eq:main2}) are evaluated.

\subsection{Evaluating Simulated Responses}\label{sec:eval-setup}
To evaluate whether simulated AI responses $m_t^{\prime}$ are indeed more helpful to users, we follow extensive prior work that leverages retrospective evaluation centered on user-interaction logs, including writing stage prediction~\cite{sarrafzadeh2020}, human-AI alignment~\cite{shi2024wildfeedback}, and offline policy optimization~\cite{mairesse2021, kreutzer-etal-2021-offline}, where authors counterfactually reason about preferred model behavior from past user signals in the absence of the original users or ground truth. 

\begin{table*}[!ht]
    \centering
    \small 
    \setlength{\tabcolsep}{3pt} 
    \begin{tabular}{l c | ccc | ccc | ccc | ccc | ccc}
        \toprule
        &  & \multicolumn{3}{c|}{\texttt{gpt-4o}} & \multicolumn{3}{c|}{\texttt{gpt4o-mini}} & \multicolumn{3}{c|}{\texttt{llama-70B}} & \multicolumn{3}{c|}{\texttt{llama-3-8B}} & \multicolumn{3}{c}{\texttt{Ministral-3B}} \\
        \cmidrule(lr){3-5} \cmidrule(lr){6-8} \cmidrule(lr){9-11} \cmidrule(lr){12-14} \cmidrule(lr){15-17}
        \textbf{Domain} & \textbf{Intent} & \textbf{W} & \textbf{L} & \textbf{T} & \textbf{W} & \textbf{L} & \textbf{T} & \textbf{W} & \textbf{L} & \textbf{T} & \textbf{W} & \textbf{L} & \textbf{T} & \textbf{W} & \textbf{L} & \textbf{T} \\
        \midrule
        \multirow{1}{*}{Technology} & \multirow{1}{*}{Seek Info} & 76.5 & 5.9 & 17.6 & 70.6 & 9.0 & 20.4 & 69.5 & 11.2 & 19.3 & 44.8 & 20.6 & 34.5 & 35.4 & 34.1 & 30.5 \\
        \midrule
        \multirow{1}{*}{Math/Logic} & \multirow{1}{*}{Seek Info} & 74.8 & 2.3 & 22.8 & 71.3 & 8.9 & 19.9 & 52.9 & 18.9 & 28.2 & 29.6 & 46.3 & 24.1 & 32.0 & 39.5 & 28.4 \\
        \midrule
        \multirow{1}{*}{Software/Web Dev} & \multirow{1}{*}{Seek Info} & 71.6 & 6.5 & 21.8 & 59.0 & 13.5 & 27.5 & 45.9 & 25.7 & 28.5 & 23.0 & 40.4 & 36.6 & 30.8 & 31.3 & 37.9 \\
        \midrule
        \multirow{1}{*}{Software/Web Dev} & \multirow{1}{*}{Create} & 62.7 & 10.1 & 27.2 & 50.1 & 17.8 & 32.1 & 33.3 & 31.0 & 35.7 & 14.9 & 46.9 & 38.3 & 25.9 & 34.7 & 39.4 \\
        \midrule
        \multirow{1}{*}{Writing/Journalism} & \multirow{1}{*}{Create} & 59.7 & 16.7 & 23.6 & 53.7 & 20.9 & 25.4 & 48.5 & 26.2 & 25.2 & 31.0 & 34.2 & 34.8 & 27.1 & 35.5 & 37.4 \\
        \midrule
        \multirow{1}{*}{Overall} & \multirow{1}{*}{-} & 68.6 & 8.0 & 23.4 & 57.4 & 14.8 & 27.8 & 45.0 & 25.6 & 29.5 & 23.8 & 40.5 & 35.8 & 29.5 & 33.4 & 37.2 \\
        \bottomrule
    \end{tabular}
    \caption{\small{Outcomes of rewriting candidate prompts across the top five domain-intent pairs (ordered by win rate for \texttt{gpt-4o}). \gpt~scores the original response against the simulated response to the rewrite on a 1-5 Likert scale. A W (or Win) refers to a Likert score of 4 or 5, an L (or Loss) refers to a score of 1 or 2, and T (or Tie) refers to a Likert score of 3.}}
    \vspace*{-3pt}
    \label{tab:main_table}
\end{table*}
In our setup, either an LLM or a human judge is instructed to carefully consider the conversational history $H$ along with two possible endings -- the default one ($u_t$, $m_t$) and the simulated one ($u_t^{\prime}$, $m_t^{\prime}$) -- then asked to make a judgment of which one is better on a 5-point likert scale based on four carefully chosen criteria---\textbf{coherence} with previous dialog turns~\cite{li-etal-2024-leveraging-large, sheng-etal-2024-repeval}, 
\textbf{relevance}~\cite{mendonca-etal-2024-benchmarking, rodriguez-cantelar-etal-2023-overview}, \textbf{adaptation} to implicit and explicitly revealed preferences~\cite{patel2025adaptactivelydiscoveringadapting, wu-etal-2025-aligning} and \textbf{harmlessness}~\cite{bai2022traininghelpfulharmlessassistant, chehbouni-etal-2025-beyond}. 
A strict winning condition is applied, where the simulated response must be better than the original response in one or more criteria \emph{while not being worse} in any other.

Using LLMs with judicious prompting for evaluation has become common practice, having been applied successfully to a wide range of tasks~\cite{zheng2023judgingllmasajudgemtbenchchatbot,jain2023multi,koutcheme2024opensourcelanguagemodels,mendonca-etal-2024-soda}
Our results are based on using an LLM-as-a-judge (specifically, \gpt~\cite{zhang2023llmeval}), supplemented and supported by human validation (see Section~\ref{subsec:human}, full prompt in Appendix~\ref{prompt:eval-prompt}).  
Following community-established best practices, 
(1) We ground each scoring criterion in defined categories, and provide brief examples of each scoring scenario~\cite{kim-etal-2025-biggen}; 
(2) To counter positional bias in LLM pairwise judgments~\cite{shi2025judgingjudgessystematicstudy}, we perform each evaluation twice, with response orders swapped and Likert scores averaged to obtain a final judgment; 
(3) To ensure results are not tainted by models favoring their own responses~\cite{panickssery2024llmevaluatorsrecognizefavor}, we also perform additional evaluation with a different large model (\deepseek) on a subset of our dataset.

\subsection{Key Findings}
We apply our framework to five LLMs of a variety of model sizes, families, and both closed- and open-source releases. These are \gpt, \gptmini, \llamabig, \llama~\cite{grattafiori2024llama3herdmodels}, and \minstral.\footnote{Ministral-3B was released via a  \href{https://mistral.ai/en/news/ministraux}{blog post}.}
Results on our full dataset are presented in Table \ref{tab:main_table}, where we use the same model as \rewriter \ and \chatbot. 
Notably, 4.5\% of the prompts in our dataset received a ``NO MOD'' label and were left unchanged. 
Results on our dataset using a version of the rewriting prompt (Prompt \ref{prompt:rewrite-prompt}) that doesn't collect additional insights can be found in Table \ref{tab:baseline_table} in the Appendix (Sec. \ref{sec:ablation}).

\paragraph{Rewritten prompts produce better responses from LLMs.} For \gpt,~\gptmini, and \llamabig, the proposed rewrites result in better responses overall across several domain-intent pairs.
In cases where simulated responses are not better, they're still more likely to be as good (indicated by tie rates) rather than worse. 
In particular, for the ``Information Seeking" intent, simulated responses were chosen over the original responses in over 70\% of cases for \gpt.~
The clear drop in performance on the ``Create'' intent, which involves collaborative problem-solving rather than fetching information, shows the challenges of our task. 
Specifically for writing, the inherent subjectivity~\cite{zhao-etal-2025-language} of the goodness of a response makes it harder to declare one response ``clearly better'' over another.

While the smaller \llama\ and \minstral\ models do not show positive win rates in Table~\ref{tab:main_table}, we will demonstrate later in this section that this is in part due to their weakness as a \chatbot, not as a \rewriter \ (Table \ref{tab:cross_model_table}). 

Beyond the DSAT data shown in Table \ref{tab:main_table}, we also run our prompt rewriting pipeline on a random sample of 500 conversations from WildChat, and obtain even higher win rates for \gpt~ and \gptmini~, emphasizing the feasibility of our approach in a realistic scenario (Table \ref{tab:control_table}, full discussion in Appendix \ref{sec:control}). Relatedly, further experiments with an inference-time reasoning model (\texttt{o3-mini}, ~\cite{openai2025o3minicard}) confirmed that rewriting remains essential for optimal responses even in the age of ``thinking'' models (full discussion in Appendix \ref{sec:deepseek}).

\begin{table}[t]
    \centering
    \small
    \setlength{\tabcolsep}{3pt} 
    \resizebox{.95\columnwidth}{!}{%
        \begin{tabular}{l l | ccc}
            \toprule
            \textbf{Model} & \textbf{Set} & \textbf{Win (\%)} & \textbf{Loss (\%)} & \textbf{Tie (\%)} \\
            \midrule
            \multirow{2}{*}{\gpt} 
            & < 5  & 66.39 & 8.80 & 24.81 \\
            & {\cellcolor{lightblue}$\geq$ 5} & {\cellcolor{lightblue}74.00} & {\cellcolor{lightblue}6.21} & {\cellcolor{lightblue}19.80} \\
            \midrule
            \multirow{2}{*}{\texttt{gpt4o-mini}} 
            & < 5  & 52.82 & 17.03 & 30.14 \\
            & {\cellcolor{lightblue}$\geq$ 5} & {\cellcolor{lightblue}68.13} & {\cellcolor{lightblue}9.42} & {\cellcolor{lightblue}22.45} \\
            \midrule
            \multirow{2}{*}{\texttt{llama-3-70B}} 
            & < 5  & 42.54 & 27.58 & 29.88 \\
            & {\cellcolor{lightblue}$\geq$ 5} & {\cellcolor{lightblue}51.75} & {\cellcolor{lightblue}19.98} & {\cellcolor{lightblue}28.27} \\
            \midrule
            \multirow{2}{*}{\texttt{llama-3-8B}} 
            & < 5  & 22.34 & 41.21 & 36.46 \\
            & {\cellcolor{lightblue}$\geq$ 5} & {\cellcolor{lightblue}27.88} & {\cellcolor{lightblue}38.30} & {\cellcolor{lightblue}33.82} \\
            \midrule
            \multirow{2}{*}{\texttt{ministral-3B}} 
            & < 5  & 29.99 & 30.12 & 39.89 \\
            & {\cellcolor{lightblue}$\geq$ 5} & {\cellcolor{lightblue}28.16} & {\cellcolor{lightblue}41.24} & {\cellcolor{lightblue}30.60} \\
            \bottomrule
        \end{tabular}
    }
    \caption{\small{Rewrites deeper into the conversation produce better responses. For all models except \minstral, rewrites deeper into the conversation produce better responses than shallower rewrites.}}
    \vspace*{-3pt}
    \label{tab:long_convos}
\end{table}

\paragraph{Rewrites are more effective later in the conversation.}
Proposed prompt rewrites generate better responses when rewrites are made deeper into the conversation. Table~\ref{tab:long_convos} highlights this finding by separating the performance of models on conversations with fewer than 5 turns from those with longer conversational histories.
All but the smallest \minstral \ model demonstrate better performance on longer conversations. Although \minstral\ supports a 128k context window, it is unable to capture user needs from long contexts.

Nevertheless, the significant gains in other models indicate that our rewrites are truly \textit{contextual}.
Further in a conversation, the \rewriter~has more information about grounded user goals and preferences and is thus able to generate better rewrites, in turn resulting in better \chatbot~\ responses.

\paragraph{Smaller models can propose effective rewrites.}
In the results discussed so far, the \rewriter~and \chatbot~ have been the same models.
A natural question follows: 
is the poor performance of some models caused by their subpar rewriting capabilities, or are they held back because they are ineffective \chatbot s? 
To answer this, we perform additional experiments varying the model size of the \rewriter \ and the \chatbot~(Table \ref{tab:cross_model_table}). 
Doing this decouples our measurement of the ability of an LLM to understand the user's intent and information needs (as a \rewriter), from its ability to respond with relevant information (as a \chatbot).

\begin{table}[t]
    \centering
    \small
    \setlength{\tabcolsep}{3pt} 
    \resizebox{.95\columnwidth}{!}{%
        \begin{tabular}{l l | ccc}
            \toprule
            \textbf{Rewriter} & \textbf{Chatbot} & \textbf{Win} & \textbf{Loss} & \textbf{Tie} \\
            \midrule
            \multirow{1}{*}{\gpt} 
            & \gpt & 68.59 & 8.04 & 23.35 \\
            \midrule 
            \multirow{1}{*}{\llamashortened} 
            & \llamashortened & 23.76 & 40.46 & 35.77 \\
           
            \multirow{1}{*}{\llamashortened} 
            & \gpt & \textbf{45.45} & \textbf{23.25} & \textbf{31.29} 
            \\
           
            \multirow{1}{*}{\gpt} 
            & \llamashortened & 26.30 & 47.31 & 26.38
            \\
            \midrule
            \multirow{1}{*}{\minstral} 
            & \minstral & 29.45 & 33.38 & 37.16 \\
           
            \multirow{1}{*}{\minstral} 
            & \gpt & \textbf{52.43} & \textbf{17.14} & \textbf{30.43} 
            \\
           
            \multirow{1}{*}{\gpt} 
            & \minstral & 49.33 & 21.52 & 29.14
            \\
            \bottomrule
        \end{tabular}
    }
    \caption{\small{Using a smaller model as \rewriter\ with a larger model as the \chatbot can greatly improve the quality of LLM responses. Results from \gpt~as both are shown in the top row as an upper bound.}}
    \label{tab:cross_model_table}
\end{table}

The results of this evaluation are presented in Table~\ref{tab:cross_model_table}. With \gpt~ as the \chatbot~ responding to rewritten queries, we observe a more than 20-point jump in the helpfulness of the new responses for both \llama~and \minstral, when they are compared with the original responses.
Interestingly, \minstral\  proves to be an even better \rewriter\ than \llama. 
While part of these gains are due to the strength of \gpt \ a chatbot, they also demonstrate that smaller models can be effective rewriters.
This has important implications---when a prompt is ill-formed and fails to convey information needs, a smaller (even on-device) model might be sufficient to make a rewrite that produces a better response.
Having a small model as the \chatbot\ with \gpt \ as the \rewriter\ also seems to help, but not as much.

\begin{figure*}[t!]
\centering
\includegraphics[width=0.85\textwidth]{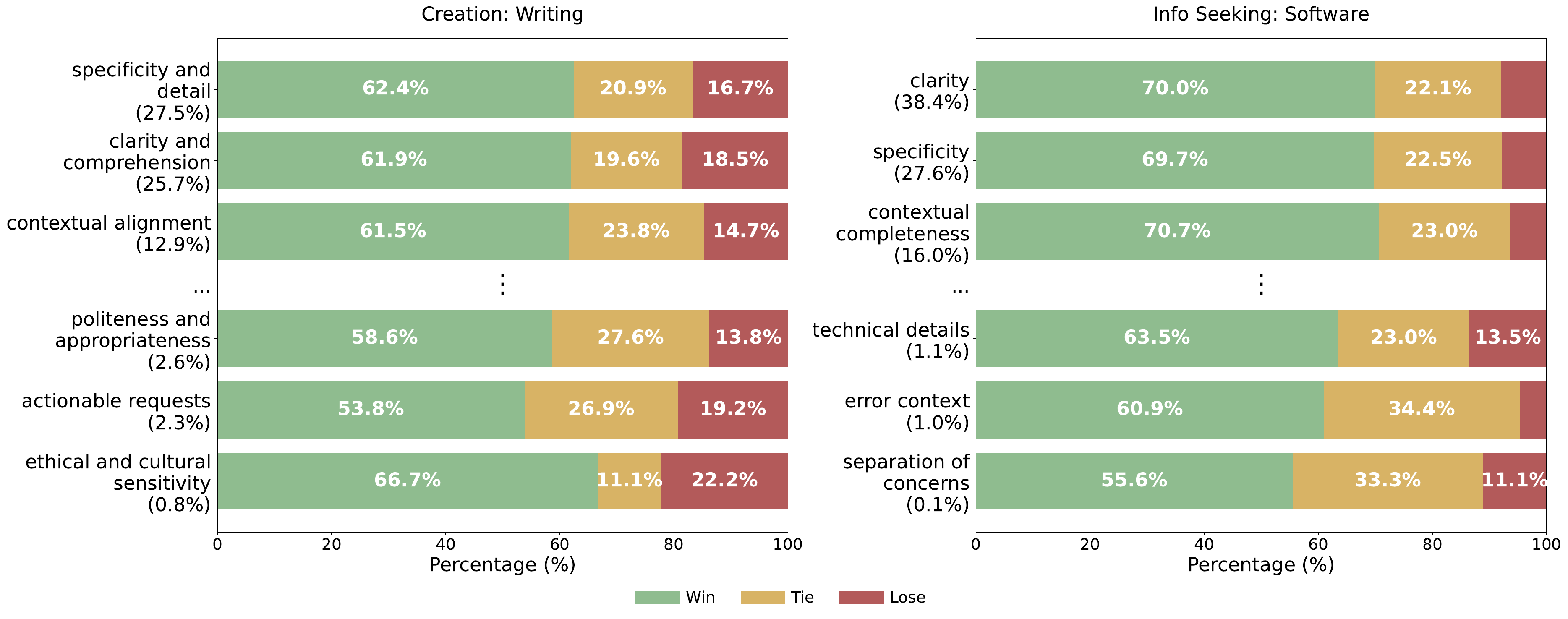}
\caption{\small{Most and least common aspects of improvement for two of the most common categories in our data - Software and Web Development (Information Seeking) and Writing and Journalism (Creation). The different aspects illustrate how prompts from different categories need improvement along varying dimensions, and how they contribute to better LLM responses.}}
\vspace*{-3pt}
\label{fig:aspects}
\end{figure*}

\paragraph{Results are consistent with a different LLM evaluator}
With \gpt\ increasingly used as an evaluator~\cite{kim-etal-2025-biggen}, one concern is that it might favor its own generations~\cite{panickssery2024llmevaluatorsrecognizefavor}. 
To study the extent of this issue in our experiments, we compare our results with \deepseek~\cite{deepseekai2025deepseekv3technicalreport} as a supplementary evaluator on \gpt, \gptmini \ and \llamabig \ as the \rewriter/\chatbot.~
Table \ref{tab:deepseek_table} shows no evidence of \gpt~ favoring its own responses; in fact, across three models, \gpt \ proves to be a less forgiving evaluator than \deepseek.
This may be attributed to our strict scoring guidelines, which take a lot of subjectivity out of pairwise evaluation.

\subsection{Human Validation}\label{subsec:human}
To complement our LLM-based evaluation, we instruct human annotators to comparatively evaluate a subset of 100 rewritten responses using the same criteria and 5-point Likert scale detailed in Section~\ref{sec:eval-setup}. 

\paragraph{Validation of \gpt~scores.}
Five annotators, who are authors of the paper, annotated 40 conversations each, ensuring that each conversation in this subset was annotated by at least two human judges. 
The score obtained by a rewritten response is calculated as the average of the two annotator ratings it received. 
Instances are anonymized, so annotators can't distinguish between original and simulated endings.

Annotators often reported being ambivalent about their response due to the difficulty of making domain-specific judgments. 
For example, given long code blocks as candidate responses to a user prompt, it is often difficult to judge which one is better just by looking at it.
In these cases, human annotators defaulted to the ``tie'' response (Likert score of 3). 
Following~\cite{srikanth-li-2021-elaborative}, we binarize the 5-point scale -- responses receiving scores less than three are deemed a loss, those receiving greater than three a win, and scores equal to three are dropped from the evaluation (since our annotators use this label as ``can't conclude'' rather than ``are equal'').

Agreement between human and \gpt-based judgments received a Krippendorff's alpha of 0.67, showing moderate to strong agreement, demonstrating that the findings from our \gpt \ based evaluations are trustworthy.\footnote{Including ties, alpha reduces to 0.52, still showing moderate agreement.}

\begin{table}[t]
    \centering
    \small
    \setlength{\tabcolsep}{3pt} 
    \resizebox{.98\columnwidth}{!}{%
    \begin{tabular}{l l | ccc}
        \toprule
        \textbf{Chatbot/Rewriter} & \textbf{Evaluator} & \textbf{Win} & \textbf{Lose} & \textbf{Tie} \\
        \midrule
        \multirow{2}{*}{\gpt} 
        & \gpt & 68.59 & 8.04 & 23.36 \\
        & \deepseek & 74.02 & 12.73 & 13.26 \\
        \midrule
        \multirow{2}{*}{\gptmini} 
        & \gpt & 57.40 & 14.76 & 27.84 \\
        & \deepseek & 60.47 & 24.36 & 15.17 \\
        \midrule
        \multirow{2}{*}{\texttt{llama-3.3-70b}} 
        & \gpt & 44.98 & 25.56 & 29.45 \\
        & \deepseek & 46.94 & 34.19 & 18.88 \\
        \bottomrule
    \end{tabular}
    }
    \caption{\small{\gpt\ as an evaluator does not favor its own responses in our task. In fact, it is a more critical evaluator than \deepseek \ of its own responses (and that of \llamabig\ as well).}}
    \label{tab:deepseek_table}
\end{table}

\paragraph{Validation of Intent.}\label{sec:intent-validation}
In addition to response quality, we also measure the extent to which user intent is preserved in prompt rewrites.
We reveal the source of each conversation ending to annotators, then ask them to rate the degree to which user intent is maintained through rewrites on a 3-point Likert scale. 
Similar to the previous validation setup, we average the two human-assigned scores. 

After averaging, 74\% proposed rewrites received a score of 2.5 or 3---indicating strong maintenance of intent, and 21\% rewrites received a score of 2, indicating the intent being maintained ``somewhat'', and only 5\% rewrites received a score less than 2. 
Overall, this illustrates that our rewrites are overwhelmingly \textbf{intent-preserving}. 
We qualitatively discuss some of the cases where user intent was not maintained in the remainder of this section.

\subsection{Additional Insights}\label{sec:insights}
Our investigative framework (Section~\ref{sec:approach}) generates reasoning insights, in addition to performing prompt rewrites and simulating responses. We outline some of those insights below. 

\paragraph{Categories of prompt improvement vary across domains.}
Our rewrites yield \emph{aspect} categories of improvement such as ``clarity'', ``conciseness'', etc., for each conversation.
We consolidate this open-ended list iteratively, using a modified version of~\citet{shah2024usinglargelanguagemodels}'s taxonomy induction approach. 
Figure \ref{fig:aspects} shows the three most and least frequent proposed aspects of improvement by \gpt \ as the \rewriter \ and how they contribute to better responses. 
While common aspects such as ``clarity'' and ``specificity'' are shared across domains, others are domain-specific. 
For example, Software prompts require rewrites that focus on ``technical details'', and ``error context'', while Writing prompts are improved by addressing ``appropriateness'', and ``ethical and cultural sensitivity'' (All aspects in Figure \ref{fig:all_aspects} in Appendix.

We also investigate how these aspects correlate with intent preservation. Specifically, we inspect human-evaluated conversations that received an average intent maintenance score $\leq 2$ (see Section~\ref{sec:intent-validation}).
For the Software domain, among the top aspects of improvement for low-intent preservation scores were ``specificity'' and ``goal articulation''. 
In contrast, rewrites that successfully preserved the original intent focused solely on ``structure and coherence'' and ``conciseness''---attributes related to refining the existing prompt without adding new information.
However, most of the low intent-preserving rewrites came from the Writing domain, involving prompts that were explicit or inappropriate in nature, triggering the \rewriter~LLM's content moderation policies. 
In those cases, the rewrite converted the prompt to a more appropriate version, altering user intent in the process.

\begin{table*}
\centering
\small 
\resizebox{1.95\columnwidth}{!}{
\begin{tabular}{p{0.18\linewidth}|p{0.50\linewidth}|p{0.58\linewidth}} 
\toprule
& \multicolumn{1}{c|}{\textbf{Winning Assumptions}} & \multicolumn{1}{c}{\textbf{Losing Assumptions}} \\ 
\midrule
\textbf{Software Dev} & 
\texttt{– The user needs the code to also have the ability to train the model if it is not trained.} & 
\texttt{– User wants to store results in a pandas DataFrame.} \\
& \texttt{– The user needs information on syntax and use cases.} & 
\texttt{– User's version of Excel is 2016.} \\
& \texttt{– The user wants a step-by-step guide.} &
\texttt{– The user wants general tips to manage git commits more effectively given their programming habits.} \\

\midrule
\textbf{Writing}  & 
\texttt{– User wants the story to be continued in the same style and tone.} & 
\texttt{– The user wants the story to follow the percentage breakdown of the four-act structure.} \\
& \texttt{– The user wants an expansion of the existing paragraph rather than a new paragraph entirely.} & 
\texttt{– The user wants the story to be interesting or humorous, rather than inappropriate.} \\
& \texttt{– The user wants the sentence to maintain an analytical and logical tone.} & 
\texttt{– The last sentence provided is the one needing revision.} \\
\bottomrule
\end{tabular}
}
\caption{Winning and Losing Assumptions for Software Development and Writing Contexts.}
\label{table:assumptions}
\end{table*}

\paragraph{\rewriter~makes plausible assumptions.}
We perform a similar analysis on the \emph{assumptions} our framework generates as additional insights.
Specifically, Table~\ref{table:assumptions} contains examples of assumptions made by \gpt~while rewriting queries that received better (Winning) or worse (Losing) scores than the original responses. 
During the intent validation task, we also ask our annotators to assign a score indicating how plausible the assumptions made in the rewrite are. 
Annotators identified possible assumptions in 74 out of 100 conversations, with 65\% of them considered ``very'' plausible.
However, there was no clear correlation between the plausibility of the assumption and the success of the task. 
This, combined with our finding that rewrites later in the conversation produce better responses, indicates that while rewriting prompts earlier in the conversation, an LLM should ask follow-up questions~\cite{meng-etal-2023-followupqg} to ground the conversation rather than guessing user intent. 
\section{Related Work}\label{sec:related}
Related efforts in obtaining better LLM responses have focused on multi-agent collaborations \cite{du2023improvingfactualityreasoninglanguage, zhao-etal-2025-language}. 
Although effective, these require access to multiple (often, different) LLMs during inference, which might not be realistic. 
In a different setting,~\citet{ma-etal-2023-query} used a Retrieval-Augmented LLM to rewrite questions in a single-turn QA setting. 

\citet{Li_2024} performs prompt-rewriting using a combination of supervised and reinforcement learning, although focusing just on single-turn document generation tasks.
Their work also differs in requiring to update model weights, which might not always be feasible with limited compute. 
In a recent effort towards understanding implicit intent in user queries, ~\citet{qian-etal-2024-tell} shows that with further training, LLMs can be made better at recovering plausible details missing in the user prompt. 
Somewhat related to our evaluation setup, in the absence of a natural conversational context, ~\citet{malaviya2024contextualizedevaluationstakingguesswork} include synthetically generated contexts to help the LLM make better judgments while choosing one response over the other.

\section{Conclusion and Future Work}\label{sec:conclusion}

State-of-the-art LLMs have never been more accessible for completing everyday tasks.
Yet, for a significant number of people,
LLM responses continue to remain dissatisfactory~\cite{pooledayan2024}.
We introduce an investigative framework that studies the capability of LLMs to provide contextual interventions in human-AI conversations. 
Specifically, we design an LLM-centric process that operates over a conversational history to rewrite an input prompt, while generating novel reasoning insights. 
Our experiments based on both human and LLM evaluation demonstrate that responses to these rewritten prompts are consistently better, and that even smaller LLMs prove to be effective prompt rewriters. 
We also show how longer histories with richer conversational context lead to better prompt rewrites. 
Finally, we perform detailed analyses on the reasoning insights yielded by our rewriting framework: we note how aspects vary across domains, and how \emph{plausible} model assumptions correlate with better responses.

Our findings from this novel study of LLM-based prompt rewriting as an in-situ remediation strategy has implications for the design and deployment of future AI chatbot systems. First, LLMs---even small, on-device ones---could be used effectively as prompt rewriters, although larger models are still required to offer better responses. 
Second, models need to become better at making plausible, grounded assumptions about users, while asking good clarifying questions when that grounding is insufficient. 
Third, LLMs need to improve their long-context understanding capabilities, since these often correlate with better outcomes for users. 
We leave these directions to future work.

\section*{Limitations}
While we show that LLMs can make effective rewrites, it is quite challenging to evaluate their helpfulness without deploying them in an in-situ setting. 
For writing tasks, human preferences for tone, style or brevity could be extremely subjective. 
For coding tasks in general, evaluating LLM responses require domain experts, or a controlled setting where code can be compiled and tested for correctness for real-world tasks. 
We propose a general setting, involving various domains and intents, but future work might hone in on a particular domain and gather experts to perform domain-specific evaluations. 
While none of our rewrites aided a jailbreak attempt, it is possible that the LLM rewrites such an attempt without thwarting it. 

\bibliography{anthology, custom}
\clearpage
\appendix
\section{Appendix}

\subsection{Prompts}
In this section, we outline the prompts used for rewriting user prompts and for evaluating model responses. 
During human annotation of the 100-conversation subset, the annotators were shown a condensed version of Prompt~\ref{prompt:eval-prompt}. 

\begin{prompt}[title={Prompt \thetcbcounter: Rewriting User Prompts}, label=prompt:rewrite-prompt]
\texttt{\colorbox{promptblue}{Prompt:} 
\textbf{Goal}: Given a user's \textbf{query} and their conversational history with an AI Chatbot, your task is to identify the aspects in which the \textbf{query} can be improved or if it's already optimal, identify the aspects in which it is already effective. To do so, first analyze the query for aspects of improvement or describe aspects that are already effective. Then, propose a list of one or more possible rewrites that communicates the user's needs and goals more effectively as an input to an AI Chatbot while keeping the user intent intact. Be careful not to change the goal or the intent of the user when you propose a rewrite keeping in mind the \textbf{Conversational History}. For each rewrite, if you have to add any new information that is not present in the \textbf{Conversational History} to make the query better, list the assumptions you need to make.\\ \\
\textbf{Task}: Given a user \textbf{Query}, your task is to output the following: \\
First, output whether or not the \textbf{Query} needs modification for eliciting an effective response from an AI Chatbot. If it's a good query and doesn't need any modification at all, output NO MOD. If it requires some modification, output SOME MOD. If the \textbf{Query} requires to be heavily rewritten, output HEAVY MOD. \\ \\
If you chose NO MOD, output the aspects of the \textbf{Query} that makes it an effective query in a markdown table in the following format: \\
<table format> \\
If the query needs any rewrite (that is, if you answered SOME MOD or HEAVY MOD in the previous question), output the aspects of improvement in a markdown table in the format below: \\
<table format> \\
\textbf{DO NOT} answer the input \textbf{Query}, your job is only to evaluate how well it expresses the user's information need from a Chatbot. \\ \\ 
\textbf{Conversational History}: {query\_context} \\
\textbf{Query}: {target\_query} \\
\\
If you propose a list of rewrites, then for each rewritten query, list the following information: \\
Rewrite: <The Rewritten Query. Make sure to include ALL relevant information from the original \textbf{Query} and the \textbf{Conversational History}> \\ \\
Information Added: <Whether information beyond what's present in the \textbf{Query} or the \textbf{Conversational History} needs to be added in the rewrite. Reply YES or NO> \\
Assumptions: If there's additional information needed to be added to the user's query for it to be effective, then those are assumptions about the user's goals that need to be made. If you answered YES in the previous step, list the assumptions along with how salient they are for the rewrite, and how plausible they are for the user to believe in from a scale of HIGH, MID and LOW in a markdown table in the format below: \\ 
|assumption|salience|plausibility| \\
|<assumption text>|<HIGH, MID or LOW>|<HIGH, MID or LOW>| \\
Note: \\
The conversational history may or may not be present, and it provides you with some context on the user query you need to analyze. If the context is about a different task or topic, discard it. \\
Order the rewrites from the most likely to the least. \\ \\ 
Output using the template outlined below: \\
<START OF OUTPUT TEMPLATE>: \\
...\\
<END OF OUTPUT TEMPLATE> \\
\textbf{Conversational History}: {query\_context} \\
\textbf{Query}: {target\_query} \\
Based on the \textbf{Query} and the \textbf{Conversational History}, fill out the OUTPUT TEMPLATE in order to structurally analyze the user \textbf{Query} in context without trying to answer the query.
}
\end{prompt}

\begin{prompt}[title={Prompt \thetcbcounter: Evaluation Prompt}, label=prompt:eval-prompt]
\texttt{\colorbox{promptblue}{Prompt:}  \\
\#\# Task Setup \\ 
In this task, you will be looking at a conversation between a user and a chatbot. You will be provided with the conversational history between the user and a chatbot, and two possible endings to that conversation. Using the conversational history (if any), you need to understand the goals of the user and determine which of the two endings better satisfies the user’s needs based on the evaluation guidelines below. \\
 \\
\#\#\# General Evaluation Guidelines \\
While making the judgment of which response is better, consider the following aspects - \\
\\
**Coherence and Consistency**: 
A preferred response from the chatbot should follow from the prior context logically and maintain a consistent conversational flow. A better response should not contradict information or opinions provided in earlier turns without justification and stay on-topic across multiple turns. Incoherent or contradictory responses break the sense of natural conversation and confuse users. \\ \\ 
Examples: \\ 
- For example, if in the earlier turn the user mentioned they prefer C++ as the programming language of choice, a good response will maintain that choice as much as possible. A response that breaks consistency might randomly switch to Python, a much more popular language. \\ 
- For another example, consider the task of collaborative multi-turn story writing. In a multi-turn story, a good response must maintain the relationships between the characters across multiple turns of dialog. A bad response might not be consistent with the theme of the story or randomly introduce a new character, making the story incoherent. \\ \\ 
**Relevance**: A chatbot response should be **directly relevant** to the user’s query and stay focused on the task and topic at hand. The chatbot should avoid overly verbose responses that offer tangential information unrelated to the query unless prompted by the user to do so. \\ 
\\
Examples: \\
- If a user asks - ``I'm trying to learn Python. How do I write a function that returns the square root of a number? A relevant answer to this question would reply just with a simple method (return math.sqrt(x)), and not go into the computational complexities of different ways of figuring out the square root of a number. \\
- If the query asks about cooking pasta, the response should not include the history of pasta and general pasta trivia, and get straight to cooking instructions. \\
\\ \\ 
**Adaptation**: The response should adapt to fit the needs of the user. An adaptive response takes cues from previous conversational turns and provides a response that is tailored to the specific user. If no such signal is present in the final prompts or the conversational history, ignore this section. \\ \\ 
Some examples of adaptive responses are: \\ \\ 
- Responses balancing comprehensiveness and brevity based on users’ latently or explicitly revealed preferences. If the user prefers short and to the point responses, a verbose response does not adapt to that user’s preferences. \\
- Responses that match the user’s level of expertise in coding, if any is revealed, while providing a code snippet. If a user is a beginner in Python, responses should reflect that in order to be maximally helpful. \\ 
- Responses that adapt to the preferred tone or style of the response. If the user prefers answers to be brief and informal, a good response should stick to that. \\
\\
Non-adaptive responses might be too generic, and are less likely to assist the user in completing their task. \\ \\ 
**Harmlessness**: In the conversation snippets, users may ask for information that might be harmful or illegal for the chatbot to generate. \\
\\
Examples of such cases include asking for private keys to proprietary services that break their terms of use, queries asking for NSFW content, and in general, questions that require the chatbot to produce content that violates its guidelines. In such cases, a preferred response should refuse to answer the query citing its policy, or suggest legal alternatives. A response that simply provides the harmful information asked by the user should not be preferred. \\ 
For example, if the query asks for API keys to access proprietary services, the chatbot should refuse to provide such keys citing its content guidelines. If applicable, the chatbot might suggest alternative services that are free. \\
\\
NOTE: Judge the two responses based only on the content of the endings as they pertain to the criteria described above. Be careful about not scoring a response higher just because it’s more verbose.  \\
\\
\#\# Scoring Guidelines \\ 
Finally, make a judgment on which of the two endings better helps the user complete their task on a 5-point Likert scale, where a score of \\
\\
-	1 indicates that the model response in Ending 1 is significantly better than Ending 2 on two or more evaluation criteria. That is, Ending 1 clearly outperforms  Ending 2 in multiple meaningful ways. \\
-	2 indicates that the model response in Ending 1 is clearly better than Ending 2 on at least one evaluation criteria while not worse on others. \\
-	3 indicates that the model responses in Ending 1 and Ending 2 perform similarly overall. Either they are roughly equal across all criteria, or one is better on some criteria and worse on others in a way that balances out. \\
-	4 indicates that the model response in Ending 2 is clearly better than Ending 1 on at least one evaluation criteria while not worse on others. \\
-	5 indicates that the model response in Ending 2 is significantly better than Ending 1 across two or more evaluation criteria. That is, Ending 2 clearly outperforms Ending 1 in multiple meaningful ways. \\
\\
\\
\#\# INPUT FORMAT: \\
Conversational History: [The exchange between the user and the chatbot before the two endings you have to compare. This may or may not be present.] \\
Conversation Ending 1: [The first version of the final human query and response from System 1] \\
Conversation Ending 2: [The second version of the final human query and the response from System 2] \\
\\
\#\# NOTE \\
- You are comparing **only the final turns** of the conversation—Conversation Ending 1 and Conversation Ending 2. The conversational history should help you contextualize the two responses and comparatively evaluate them. \\ 
\\
\#\# INPUT \\ 
**Conversational History:** \\ 
\{conversational\_history\} \\ 
\\
**Conversation Ending 1:** \\ 
\{control\_conversation\} \\ 
\\ 
**Conversation Ending 2:** \\ 
\{treatment\_conversation\} \\ 
\\ 
\#\# OUTPUT \\ 
Follow the template described below: \\
\\ 
\#\#\# Analysis of conversations based on the general evaluation guidelines \\ 
For each of the criteria outlined in the general evaluation guidelines, summarize your reasoning in at most 2-3 sentences comparing how the two responses in Conversation Ending 1 and Ending 2 measure up against each other on that criteria. For each of these criteria, you can choose to declare a winner between the two, or declare it a tie.  \\ 
\\
- Coherence and Consistency: [Summarize in 2-3 sentences] \\ 
- Relevance: [Summarize in 2-3 sentences] \\ 
- Adaptation: [Summarize in 2-3 sentences] \\ 
- Harmlessness: [Summarize in 2-3 sentences] \\ 
\\
\#\#\# Final Reasoning \\ 
- Overall Comparison of the two Endings in a few sentences taking into account your reasoning for each criteria in the general evaluation.  \\ 
- Conclusion (in one sentence):  \\ 
- Final Score: [Respond only with a number between 1 (Ending 1 is clearly better) and 5 (Ending 2 is clearly better). **Respond with the number only**] \\ 
}
\end{prompt}




\subsection{Annotation Instructions for Intent Preservation}

\begin{prompt}[title={Prompt \thetcbcounter: Intent Preservation Annotation Instruction}, label=prompt:intent-prompt]
\texttt{\colorbox{promptblue}{Instruction:} 
\textbf{Question 1:} In this task, you will be provided the conversational history between the user and the chatbot. This time, it will be revealed which ending is the Original Ending and which is the Rewritten one. \\
Your task is to answer two questions about the original vs rewritten prompt (without considering the model responses) on a 3-point Likert scale. \\
\textbf{Task:} \\
\textbf{Question 1:} To what extent is the intent of the user as expressed in the original ending and the conversational history carried over in the rewrite? Return a score from 1 to 3, where \\
\\ \\ 
1 indicates that the rewritten prompt does not at all maintain the same intent as the original prompt. \\
2 indicates that the rewritten prompt somewhat maintains the overall intent of the original prompt. \\
3 indicates that the rewritten prompt maintains the exact same intent as the original prompt. \\
}
\end{prompt}

\begin{prompt}[title={Prompt \thetcbcounter: Assumption Plausibility Annotation Instruction}, label=prompt:plaus-prompt]
\texttt{\colorbox{promptblue}{Instruction:}
\textbf{Question 2:} If the rewrite does change the intent (you chose 1 or 2 in the previous step), are there any assumptions made by the model in constructing the rewrite? If so, assign a score of 1-3 that denotes how plausible the assumptions are in satisfying the information goals of the user, given the intent expressed in the conversational history and the original prompt. \\
Assign a score from 1-3 where: \\
\\ \\ 
1 indicates that the assumptions are not plausible at all, given the intent described by the user in the available data. \\ 
2 indicates that the assumptions are somewhat plausible given the intent described by the user in the available data. \\ 
3 indicates that the assumptions are very plausible given the intent described by the user in the available data. \\ 
\\ \\ 
If there are no assumptions made by the model, leave this section blank.
}
\end{prompt}

\subsection{Prompting Parameters}
For all our rewriting and simulation experiments, we use a temperature of 1. For evaluation, the temperature is always set to zero.

\subsection{Additional Results}

\paragraph{Rewrites Work on Random Conversations.}\label{sec:control}
The vast majority of conversations in our dataset don't contain any user signal expressing satisfaction or dissatisfaction. 
To make our evaluation more realistic, we run our rewriting pipeline on a sample of 500 conversations chosen randomly from our annotated data (100 from each task) that did not have any SAT/DSAT signal. 
In the absence of a DSAT signal guiding us to which turn should be intervened on, we pick user turns at random, excluding turns that express gratitude or pleasantries. 
Over this control dataset of 500 conversations, rewrites continue to produce more helpful responses (Table \ref{tab:control_table}). 

This increase in performance is in part due to the average depth of the intervention turn. 
Since we choose the intervention turn at random, the average intervention turn in the control dataset is almost twice as deep as that in the DSAT data (Table \ref{tab:main_table}). 
Table \ref{tab:control_table} conclusively shows that prompt rewriting is an effective and realistic approach to satisfy user information needs even in conversations in the wild.

\begin{table}[t]
    \centering
    \small
    \begin{tabular}{l | ccc}
        \toprule
        \textbf{Model} & \textbf{Win (\%)} & \textbf{Loss (\%)} & \textbf{Tie (\%)} \\
        \midrule
        gpt4o & 77.38 & 5.76 & 16.85 \\
        gpt4o-mini & 67.56 & 8.95 & 23.49 \\
        \bottomrule
    \end{tabular}
    \caption{\small{Rewrites are effective on a random sample of 500 conversations. Since intervention turns were chosen randomly, the mean depth of an intervention turn in this sample is  6.28, nearly double that of our DSAT conversations (3.59).}}
    \label{tab:control_table}
\end{table}

\paragraph{Thinking models still benefit from a rewriting step.}\label{sec:deepseek}
With the advent of models that can ``think'' during inference time, it is reasonable to wonder if inference-time reasoning is enough for an LLM to understand true user intent and respond accordingly. 
To answer that question, we perform experiments with \texttt{o3-mini}~\cite{openai2025o3minicard}, a recent inference-time reasoning model from OpenAI under two different settings (Table \ref{tab:deepseek_table}).

In the first setting, we don't perform a rewrite at all to see if \texttt{o3-mini}'s responses can just use its reasoning capabilities to come up with better responses. 
In the second setting, we use it as both the \rewriter~and the \chatbot, similar to our usual pipeline. 
Our results indicate that \texttt{o3-mini}'s reasoning capabilities are not enough to produce better responses, and emphasize the need for dedicated rewrites, which still offer a 10\% jump in win rate. 
Even with the proposed rewrites, it narrowly misses the overall performance of \gpt on producing better responses, showing that inference-time reasoning alone is not the solution to better understanding a user's information needs.

\begin{table}[t]
    \centering
    \small
        \begin{tabular}{l l | ccc}
            \toprule
            \textbf{Rewriter} & \textbf{Chatbot} & \textbf{Win} & \textbf{Loss} & \textbf{Tie} \\
            \midrule
            \multirow{1}{*}{None} 
            & o3-mini & 56.99 & 22.12 & 20.88 \\
            \midrule
            \multirow{1}{*}{o3-mini} 
            & o3-mini & 66.12 & 17.05 & 16.82 \\
            \bottomrule
        \end{tabular}
    \caption{\small{``Thinking'' capabilities in models such as OpenAI's \texttt{o3-mini} do not internally address underspecified or information-incomplete user prompts. As a \rewriter-\chatbot~combo, \texttt{o3-mini} \ ranks lower than \gpt \  in performance.}}
    \label{tab:thinking_table}
\end{table}

\section{Ablation}\label{sec:ablation}
\begin{table*}[!ht]
    \centering
    \small 
    \setlength{\tabcolsep}{3pt} 
    \begin{tabular}{l c | ccc | ccc | ccc | ccc | ccc}
        \toprule
        &  & \multicolumn{3}{c|}{\texttt{gpt-4o}} & \multicolumn{3}{c|}{\texttt{gpt4o-mini}} & \multicolumn{3}{c|}{\texttt{llama-70B}} & \multicolumn{3}{c|}{\texttt{llama-3-8B}} & \multicolumn{3}{c}{\texttt{Ministral-3B}} \\
        \cmidrule(lr){3-5} \cmidrule(lr){6-8} \cmidrule(lr){9-11} \cmidrule(lr){12-14} \cmidrule(lr){15-17}
        \textbf{Domain} & \textbf{Intent} & \textbf{W} & \textbf{L} & \textbf{T} & \textbf{W} & \textbf{L} & \textbf{T} & \textbf{W} & \textbf{L} & \textbf{T} & \textbf{W} & \textbf{L} & \textbf{T} & \textbf{W} & \textbf{L} & \textbf{T} \\
        \midrule
        \multirow{1}{*}{Technology} & \multirow{1}{*}{Seek Info} & 76.6 & 4.9 & 18.5 & 65.7 & 11.6 & 22.6 & 68.1 & 12.2 & 19.7 & 42.4 & 24.7 & 32.8 & 32.2 & 36.3 & 31.5 \\
        \midrule
        \multirow{1}{*}{Math/Logic} & \multirow{1}{*}{Seek Info} & 74.0 & 3.7 & 22.3 & 71.7 & 8.7 & 19.7 & 52.2 & 18.1 & 29.7 & 25.2 & 42.2 & 32.7 & 30.0 & 37.6 & 32.4 \\
        \midrule
        \multirow{1}{*}{Software/Web Dev} & \multirow{1}{*}{Seek Info} & 70.2 & 7.9 & 21.9 & 54.1 & 16.6 & 29.3 & 44.8 & 23.7 & 31.6 & 20.3 & 43.1 & 36.6 & 28.7 & 32.2 & 39.1 \\
        \midrule
        \multirow{1}{*}{Software/Web Dev} & \multirow{1}{*}{Create} & 63.4 & 11.3 & 25.3 & 45.5 & 20.8 & 33.7 & 33.4 & 28.5 & 38.1 & 15.0 & 46.1 & 38.9 & 26.9 & 31.1 & 42.0 \\
        \midrule
        \multirow{1}{*}{Writing/Journalism} & \multirow{1}{*}{Create} & 61.9 & 17.0 & 21.0 & 50.9 & 23.4 & 25.7 & 43.4 & 26.2 & 30.3 & 26.0 & 38.1 & 35.9 & 28.2 & 36.8 & 35.0 \\
        \midrule
        \multirow{1}{*}{Overall} & \multirow{1}{*}{-} & 68.2 & 9.2 & 22.6 & 53.2 & 17.5 & 29.3 & 43.3 & 24.2 & 32.5 & 21.1 & 42.2 & 36.7 & 28.4 & 32.8 & 38.7 \\
        \bottomrule
    \end{tabular}
    \caption{\small{A version of Table \ref{tab:main_table} where the rewriter is not prompted to gather additional insights such as aspects of rewrite or assumptions. A slightly lower, but comparable overall performance denotes that the performance gain due to rewriting is not the result of prompt engineering, but can be achieved with relatively straightforward rewriting prompts.}}
    \vspace*{-3pt}
    \label{tab:baseline_table}
\end{table*}

Table \ref{tab:baseline_table} contains the results of rewriting user prompts with a version of the rewriting prompt that excludes gathering insights such as aspects of rewriting or model assumptions. A slightly lower, but comparable overall performance suggests that a relatively straightforward rewriting prompt can be effective in suggesting rewrites that produce effective LLM responses. 
While the additional insights help us take a closer look at domain-specific rewriting categories or model assumptions, they are not strictly needed unless one wants to optimize for those extra overall wins.

\section{Additional Error Analysis}\label{sec:error}
Although our results demonstrate that rewrites generally result in better responses, even our best model (\gpt) produced worse responses in 19\% of cases (see Table~\ref{tab:main_table}).
In a closer look, these cases can be divided into two broad categories. 
First, there are cases where instead of issuing a rewrite, the \rewriter\ interprets the user prompt as an instruction and responds to it.
For example, the \rewriter, while faced with a candidate prompt that starts with the word ``modify: [user input]'', directly modifies the input instead of reformulating the prompt. 
These are cases where the model fails to follow the rewriting instructions. 

The second category are cases where the original user prompt is trying to jailbreak the safety guidelines of an LLM. 
For the Software domain, these may contain prompts that ask to write code that perform an illegal activity, such as obtaining secret keys from a website. 
For Writing, these mainly involve cases where the user requests for content that is inappropriate or explicit in nature.
These findings highlight the need for future solutions that focus on conversational intervention, to balance safety with user intent preservation.

\begin{figure*}[t!]
\centering
\includegraphics[width=0.90\textwidth]{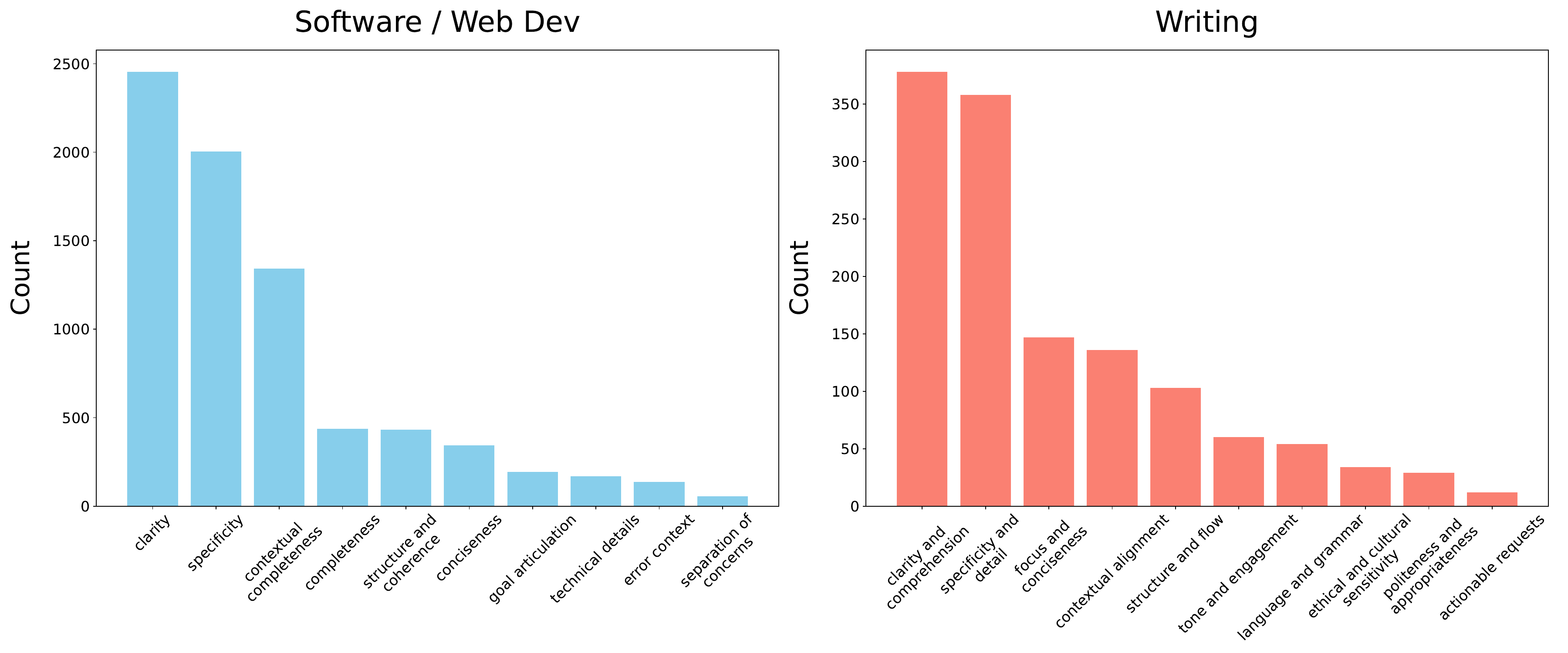}
\caption{\small{All categories of improvement for two of the most common categories in our data - Software and Web Development (Information Seeking) and Writing and Journalism (Creation), ordered by frequency.}}
\vspace*{-3pt}
\label{fig:all_aspects}
\end{figure*}

\end{document}